\documentclass{article}
\usepackage{ijcai26}

\usepackage{times}
\usepackage{soul}
\usepackage{url}
\usepackage[hidelinks]{hyperref}
\usepackage[utf8]{inputenc}
\usepackage[small]{caption}
\usepackage{graphicx}
\usepackage{amsmath}
\usepackage{amsthm}
\usepackage{amssymb}
\usepackage{booktabs}
\usepackage{multirow}
\usepackage[utf8]{inputenc}
\usepackage{xcolor}
\usepackage[linesnumbered,ruled,vlined]{algorithm2e}
\usepackage[switch]{lineno}
\usepackage{float}

\SetKwInput{KwInput}{\textsc{Input}}
\SetKwInput{KwOutput}{\textsc{Output}}
\SetKw{KwRet}{return}

\SetCommentSty{mycommfont}
\SetKwProg{Fn}{\textsc{Function}}{:}{}

\title{Post Hoc Extraction of Pareto Fronts for Continuous Control}

\author{
Raghav Thakar$^{1}$
\and
Gaurav Dixit$^{1}$\And
Kagan Tumer$^1$\\
\affiliations
$^1$The Collaborative Robotics and Intelligent Systems (CoRIS) Institute
\\Oregon State University, Corvallis, Oregon, USA\\
\emails
\{thakarr, dixitg, kagan.tumer\}@oregonstate.edu
}

\begin{document}

\maketitle

\begin{abstract}
Agents in the real world must often balance multiple objectives, such as speed, stability, and energy efficiency in continuous control.
To account for changing conditions and preferences, an agent must ideally learn a Pareto frontier of policies representing multiple optimal trade-offs.
Recent advances in multi-policy multi-objective reinforcement learning (MORL) enable learning a Pareto front directly, but require full multi-objective consideration from the start of training.
In practice, multi-objective preferences often arise after a policy has already been trained on a single specialised objective.
Existing MORL methods cannot leverage these pre-trained `specialists' to learn Pareto fronts and avoid incurring the sample costs of retraining.
We introduce Mixed Advantage Pareto Extraction (MAPEX), an offline MORL method that constructs a frontier of policies by reusing pre-trained specialist policies, critics, and replay buffers.
MAPEX combines evaluations from specialist critics into a mixed advantage signal, and weights a behaviour cloning loss with it to train new policies that balance multiple objectives.
MAPEX's post hoc Pareto front extraction preserves the simplicity of single-objective off-policy RL, and avoids retrofitting these algorithms into complex MORL frameworks.
We formally describe the MAPEX procedure and evaluate MAPEX on five multi-objective MuJoCo environments.
Given the same starting policies, MAPEX produces comparable fronts at $0.001\%$ the sample cost of established baselines.
\end{abstract}

\section{Introduction}

Many real-world continuous control problems require agents to balance multiple, even conflicting, objectives.
In legged locomotion, for example, a robot must simultaneously optimise forward speed, gait stability, and energy efficiency.
In such settings, there is no single optimal solution, and agents must instead discover a set of non-dominated policies that capture the range of feasible trade-offs.
Learning this \emph{Pareto front} enables downstream stakeholders to select behaviours that reflect their current preferences or adapt to changing operational conditions.

Typically, multiple objectives are handled through scalarisation into a single weighted sum and training via standard reinforcement learning (RL).
While accessible and tractable, this approach yields only a single, fixed trade-off.
Simply repeating this procedure for every trade-off is inefficient and practically infeasible without sophisticated experience and representation sharing.

Multi-Objective RL (MORL), and specifically \textit{multi-policy} MORL algorithms partially address this through various architectural improvements to learn the entire frontier directly.
Methods like MORL/D~\cite{morld} divide the problem into several single-objective problems through scalarisation, and use single-objective RL with buffer sharing to train policies for each scalarisation.
Other works like PG-MORL~\cite{pgmorl} and MOPDERL~\cite{mopderl} maintain populations of policies, combining RL with evolutionary principles to jointly improve the policies in the population and cover the objective space.
PSL-MORL~\cite{psl-hypernetworks-morl}, a notable recent work, trains a hypernetwork to produce a fresh policy for every desired trade-off.

Despite successfully learning Pareto fronts, these methods suffer from a critical limitation: requiring full multi-objective consideration from the outset of training.
This rigidity creates a disconnect with practical scenarios, where multi-objective preferences often arise retroactively, typically only after a robust policy for a primary task has already been trained.
For instance, a preference for more stability may only emerge after observing a locomotion policy highly optimised for speed.
To obtain new trade-offs in these scenarios, practitioners must 1) discard pre-trained policies and incur the sample costs of re-training, and 2) retrofit their algorithm into complex multi-objective learning frameworks.
Currently, no method reuses disjoint specialist policies and training data to recover Pareto fronts efficiently and with minimal added algorithmic complexity.

In this work we present Mixed Advantage Pareto Extraction (MAPEX)\footnote{Code: https://github.com/raghavthakar/MAPEX}, a novel method that fully leverages prior-trained single-objective policies, critics, and replay buffers to produce Pareto fronts of policies.
Our key insight is that agents learn optimal trade-offs by intelligently blending expert behaviour on each objective. 
MAPEX implements this by blending the single-objective evaluation from each specialist critic into a multi-objective \textit{mixed advantage} value and weighting a behaviour cloning loss with it.

MAPEX bypasses the need to retrofit bespoke or standard off-policy RL into complex multi-objective learning frameworks.
It preserves both, the simplicity, and intricacies of these algorithms by providing a realistic pathway to learning multi-objective behaviours from single-objective training data.
If training from scratch, MAPEX allows allocating the entire sample budget to training high-performing single-objective specialists, from which high-quality Pareto fronts can later be extracted at a minimal sample cost.

We provide a detailed view of MAPEX's Pareto extraction procedure, and perform Pareto extraction from specialists trained using both standard and bespoke off-policy RL.
Across five multi-objective MuJoCo environments, MAPEX produces fronts comparable to, or better than established baselines from the literature.
When extracting Pareto fronts from the same starting policies, other methods consume up to 1000$\times$ more samples than MAPEX to produce similar fronts.

\section{Background}
\label{sec:background}
\subsection{Actor--Critic Methods and Offline Reinforcement Learning}

\paragraph{Actor--critic methods:}These methods perform reinforcement learning using an \emph{actor}, which defines the policy, and a \emph{critic}, which estimates the expected return of the actor’s behaviour \cite{sutton-barto-reinforcement-learning-introduction}.
By relying on learned value estimates rather than Monte Carlo returns alone, actor--critic methods enable efficient and stable policy optimisation, and are commonly used in continuous control.
These methods are often implemented in an \emph{off-policy} setting, where experience collected by one or more behaviour policies is stored in a replay buffer and reused for policy updates \cite{soft-actor-critic,twin-delayed-deep-deterministic-policy-gradient}.

Policy improvement is commonly guided by the \emph{advantage} function,
\begin{equation}
A^\pi(s,a) = Q^\pi(s,a) - V^\pi(s),
\end{equation}
where $Q^\pi(s,a)$ is the critic's action-value function, and $V^\pi(s) = \mathbb{E}_{a\sim\pi(\cdot|s)}[Q^\pi(s,a)]$.
In practice, particularly in deterministic actor--critic methods, the value function is often approximated by $Q^\pi(s,\pi(s))$, yielding
\begin{equation}
A^\pi(s,a) \approx Q^\pi(s,a) - Q^\pi(s,\pi(s)).
\end{equation}

\paragraph{Offline Reinforcement Learning:}
Offline reinforcement learning studies the problem of learning a policy from a fixed dataset of transitions, without any further interaction with the environment during training.
A key challenge in this setting is distributional shift: the dataset may contain diverse experiences that are hard to generalise to, leading to unreliable value estimates \cite{distribution-shift-offline-rl,offline-rl-tutorial-perspective}.

Advantage-weighted regression (AWR) \cite{awr} provides a simple and stable mechanism for policy improvement in this setting by casting reinforcement learning as a weighted supervised learning problem.
Given a critic-derived advantage estimate $A(s,a)$, the AWR policy update solves
\begin{equation}
\pi^{+}
=
\arg\max_{\pi}
\;\mathbb{E}_{(s,a)\sim \mathcal{D}}
\left[
\log \pi(a|s)\;
\exp\!\left(\tfrac{1}{\beta} A(s,a)\right)
\right],
\end{equation}
where $\beta > 0$ is a temperature parameter.
This objective emphasises actions that are predicted to improve performance while sampled purely from the observed data, making it well suited to off-policy and offline reinforcement learning problems.
We use an AWR-inspired regression weighting in MAPEX to learn multi-objective behaviours from static single-objective replay buffers.

\subsection{Multi-Objective Sequential Decision-Making}

Multi-objective sequential decision-making problems are commonly modelled as Multi-Objective Markov Decision Processes (MOMDPs).
A MOMDP is defined by the tuple
$\langle \mathcal{S}, \mathcal{A}, \mathcal{T}, \gamma, \mathbf{R} \rangle$,
containing the state space $\mathcal{S}$, the action space $\mathcal{A}$, the transition dynamics $\mathcal{T}: \mathcal{S} \times \mathcal{A} \times \mathcal{S} \rightarrow [0,1]$, and a scalar discount factor $\gamma \in [0,1)$.
Unlike standard MDPs, the reward function
$\mathbf{R}: \mathcal{S} \times \mathcal{A} \times \mathcal{S} \rightarrow \mathbb{R}^k$
returns a vector of $k$ rewards, each corresponding to a distinct objective.

A policy $\pi: \mathcal{S} \rightarrow \mathcal{A}$ induces an expected return for each objective.
We characterise policy performance directly through its expected long-term returns.
Let $\mathbf{J}(\pi) \in \mathbb{R}^k$ denote the vector of objective returns achieved by policy $\pi$, with components
\begin{equation}
J_i(\pi)
=
\mathbb{E}_{s_0\sim\rho_0,\; a_t\sim\pi(\cdot\mid s_t)}
\!\left[\sum_{t=0}^{\infty}\gamma^t r_i(s_t,a_t)\right],
\; i\in\{1,\dots,k\}.
\end{equation}

where $r_i$ is the reward function associated with the $i$-th objective and $\rho_0$ is the initial state distribution.

From a multi-objective optimisation perspective, learning in a MOMDP can be viewed as maximising a vector-valued objective
\begin{equation}
\max_{\pi}\; \mathbf{F}(\pi)
\;\triangleq\;
\max_{\pi}\; [J_1(\pi), J_2(\pi), \dots, J_k(\pi)].
\end{equation}

\paragraph{Pareto Dominance and Optimality}
As objectives are often conflicting, it is generally impossible for a single policy to optimise all objectives simultaneously.
Thus, optimality in multi-objective settings is defined in terms of \emph{Pareto dominance}.
Given two objective return vectors $\mathbf{v}, \mathbf{u} \in \mathbb{R}^k$, $\mathbf{v}$ \emph{Pareto-dominates} $\mathbf{u}$ (denoted $\mathbf{v} \succ_p \mathbf{u}$) if $\mathbf{v}$ is at least as good as $\mathbf{u}$ in all objectives and strictly better in at least one:
\begin{equation}
\mathbf{v} \succ_p \mathbf{u}
\iff
(\forall i: v_i \ge u_i) \land (\exists j: v_j > u_j).
\end{equation}
If neither vector Pareto-dominates the other, they are said to be \emph{non-dominated}.

A policy $\pi$ is \emph{Pareto optimal} if its return vector $\mathbf{J}(\pi)$ is not Pareto-dominated by that of any other policy.
The set of all Pareto-optimal policies induces the \emph{Pareto front}, which characterises the optimal trade-offs achievable in the objective space.

\section{Related Works}
Recent surveys \cite{practical-guide-to-morl,multi-objective-sequential-decision-making} categorise the expanding MORL landscape into single-policy and multi-policy approaches. We adopt this taxonomy herein.

Single-policy methods optimise a fixed scalar utility function \cite{practical-guide-to-morl}, using standard RL with linear utilities, and dedicated approaches for non-linear utilities \cite{smooth-tchebycheff-scalarization,scalar-morl-chebycheff,actor-critic-morl-nonlinear}. Conversely, multi-policy approaches approximate the Pareto front, a better-suited approach for unknown or dynamic preferences. Early works like Pareto Q-Learning \cite{pareto-q-learning} tracked non-dominated return vectors for each state-action pair, while recent extensions employ preference-conditioned deep networks \cite{dynamic-weights-morl} or convex envelope updates \cite{envelope-q-learning} for high-dimensional spaces.

In continuous control, decomposition-based methods divide the problem into several scalarised objectives. Prior work \cite{multi-policy-sac-es} has augmented naively solving each scalar objective with an evolutionary strategy \cite{evolution-strategies} for post-processing.
Notably, MORL/D \cite{morld} improves efficiency via cooperative buffer sharing and intelligent weight sampling.
Another branch of work combines RL with evolutionary operators: PG-MORL \cite{pgmorl} iteratively pushes a population of policies toward promising objective space regions, while MOPDERL \cite{mopderl} distills a frontier from subpopulations trained via evolutionary RL \cite{pderl,erl}.
Alternatively, parameter-efficient methods consolidate policies using meta-learning \cite{meta-morl}, universal preference-conditioned networks \cite{pdmorl}, or hypernetworks \cite{psl-hypernetworks-morl}.

Despite architectural differences, these methods share a structural limitation: they are designed solely for ``from scratch" learning.
If training on one or more individual objectives has already been performed, these methods cannot leverage it, effectively requiring pre-trained policies to be discarded and training to be restarted.

A separate line of inquiry focuses on offline MORL, training preference-conditioned agents from massive datasets.
Approaches like PEDA \cite{peda}, DiffMORL \cite{diffmorl}, and PR-MORL \cite{prmorl} utilise transformers, diffusion, and regularisation to generalise across preferences.
While seemingly similar, MAPEX tackles the distinct problem of extracting frontiers from pre-trained specialists rather than large-scale generalisation. Consequently, MAPEX operates with replay buffers two orders of magnitude smaller than the D4MORL benchmarks \cite{peda} used in offline RL works (e.g., 1 million vs. 150 million transitions).

Finally, PCN \cite{pcn} also uses supervised learning from an off-policy dataset, but relies on iterative online data collection to populate the objective space.
In contrast, MAPEX addresses the strictly offline extraction of frontiers from fixed, disjoint datasets.
PCN is also currently limited to discrete action spaces.

\section{Method}
\label{sec:method}

\begin{algorithm}[h]
\caption{Mixed Advantage Pareto Extraction (MAPEX)}
\label{alg:MAPEX}
\DontPrintSemicolon

\KwInput{Policy set $\Pi$, critic set $\mathcal{Q}$, buffer set $\mathcal{D}$, number of objectives $N$}
\KwOutput{Pareto front $\mathcal{P}$ of policies}

\Fn{\textsc{MAPEX}($\Pi, \mathcal{Q}, \mathcal{D}, N$)}{
    \While{Required}{
        \textsc{Evaluate}($\Pi$)\;
        \BlankLine
        $\mathcal{P} \gets \textsc{FindParetoFront}(\Pi)$\;
        \BlankLine
        $\{ \pi_1, \dots, \pi_N \} \gets \textsc{SelectParents}(\mathcal{P})$\;
        \BlankLine
        $\mathbf{v} \gets \textsc{Centroid}(\pi_1, \dots, \pi_N)$\;\tcp{In objective space}
        \BlankLine
        $\mathbf{w}_{\text{target}} \gets \frac{\mathbf{v}}{\| \mathbf{v} \|_2}\;$\tcp{Target weight vector}
        \BlankLine
        $\mathcal{D}_{\text{hybrid}} \gets \bigcup_{k=1}^N \textsc{Sample}(\mathcal{D}_k, \propto w_k)$\;
        \BlankLine
        $\pi_{\text{new}} \gets \textsc{InitPolicy}(\emptyset)\;$
        \BlankLine
        \For{$\epsilon \gets 1$ \KwTo $E$}
        {
            \BlankLine
            $(s,a)\sim\mathcal{D}_{\text{hybrid}}$\;
            \BlankLine
            $\mathbf{A} \gets \bigg[\Big( \mathcal{Q}_i(s,a) - \mathcal{Q}_i\big(s, \pi_{\text{new}}(s)\big) \Big)\bigg]_{i=1}^N$\;
            \BlankLine
            $A_{\text{mixed}} \gets \mathbf{w}_{\text{target}}^\top \mathbf{A}\;$\tcp{Mixed advantage}
            \BlankLine
            $\omega(s,a) = \text{min}\left( \exp\left( \frac{A_{\text{mixed}}(s, a)}{\beta} \right), \omega_{max} \right)$
            \BlankLine
            $\mathcal{L}_{\text{MAPEX}} \gets \mathbb{E} \Big[ \omega(s,a) \cdot \| a - \pi_{\text{new}}(s) \|_2^2 \Big]$
            \BlankLine
            $\textsc{Update}(\pi_{\text{new}}, \mathcal{L}_{\text{MAPEX}})\;$
        }
        $\Pi\gets\Pi\cup\{\pi_{\text{new}}\}$
    }
    \BlankLine
    $\mathcal{P} \gets \textsc{FindParetoFront}(\Pi)$\;
    \BlankLine
    \KwRet $\mathcal{P}$\;
}
\end{algorithm}

We now introduce Mixed Advantage Pareto Extraction (MAPEX), an offline algorithm to extract a Pareto front of continuous control policies from a set of disjoint \textit{single-objective specialists}.
The core intent of MAPEX is to fully leverage the latent information in prior-trained specialists---which includes their policies, along with their value functions (critics) and replay buffers---to produce new policies that express new trade-off behaviours without requiring additional training interaction with the environment.

MAPEX achieves this via a Pareto front extraction procedure which iteratively fills gaps in the Pareto front estimate.
First, MAPEX analyses the available policies in the objective space to identify sparse regions in the Pareto front estimate.
Once a gap is identified, MAPEX derives a vector of `target weights' that encodes a weighting over the objectives that would optimally fill this gap.
To train a policy for this new trade-off, MAPEX constructs a static \textit{hybrid buffer} by sampling from the specialists' buffers in proportion to these target weights.
Crucially, the algorithm then calculates a \textit{mixed advantage} for every transition in this dataset by querying each specialist critic, and mixing the individual advantage estimates in the ratio of the target weights.
This mixed advantage captures the value of a transition in demonstrating target trade-off behaviour.
Finally, a fresh policy is trained using a mechanism inspired by Advantage Weighted Regression (AWR) \cite{awr}, wherein we regress the policy onto actions from the hybrid buffer, weighted by an exponential of their calculated mixed advantage.
Algorithm \ref{alg:MAPEX} provides a more rigorous look at this procedure.

\subsection*{Starting Information and Notation}
We assume that from prior training we have a policy set $\Pi$ with at least $N$ policies for an $N$-objective problem, a critic set $\mathcal{Q}$ with $N$ critics, each specialising on evaluating on one of the problem's objectives, and a replay buffer set $\mathcal{D}$ with $N$ buffers, each containing experiences from training on a single objective.

\subsection{Step 1: Gap Identification and Parent Selection}
\label{step:1}
At the start of each iteration, we evaluate the current policy set $\Pi$ to obtain the performance vector $J(\pi) \in \mathbb{R}^N$ of each policy.
We then identify the non-dominated set (the current Pareto front approximation).
We then search for the largest $N$-dimensional `gap' on the frontier---a sparse region in the objective space.
For $N=2$, we define this as the edge with the maximal Euclidean span.
In practice, to avoid repeatedly focusing on the same gap, we select the gap using roulette-wheel sampling based on edge length.
The $N$ policies corresponding to the vertices of this gap are selected as the \textit{parent policies}, denoted by $\{\pi_{p_1}, \dots, \pi_{p_N}\}$.

To guide the offspring into this sparse region, we compute the centroid of the parents' performance vectors in the objective space:
\begin{equation}
    J_{mid} = \frac{1}{N} \sum_{i=1}^{N} J(\pi_{p_i})
\end{equation}
We then derive a unit vector $\mathbf{w}_{\text{target}}$ of target weights pointing towards this centroid.
This vector encodes the linear preference required to interpolate the gap and guides the subsequent hybrid buffer creation and advantage mixing steps.

\subsection{Step 2: Hybrid Buffer Creation and Advantage Mixing}
\label{step:2}
After deriving the target weight vector $\mathbf{w}_{\text{target}}$, we assemble a training distribution that reflects this desired trade-off.
We construct a fixed-size \textit{hybrid buffer}, $\mathcal{D}_{\text{hybrid}}$, by sampling transitions from each specialist's buffer $\mathcal{D}_k$ in direct proportion to the corresponding weight $\textbf{w}_{\text{target}, k}$.
This creates a dataset that is structurally biased to each objective in the desired proportion.

We then initialise a random policy network $\pi_{\text{new}}$ that will be optimised to achieve the target trade-off behaviour.
For this optimisation, we iteratively sample transitions from $\mathcal{D}_{hybrid}$ to compute the mixed advantage training signal.
For each transition $(s, a)$, we compute a vector of advantages $\mathbf{A}(s,a) \in \mathbb{R}^N$.
The $k^{th}$ element of this advantage vector is the advantage on the $k^{th}$ objective associated with that transition, and is computed using the $k^{th}$ specialist critic $\mathcal{Q}_k$:
\begin{equation}
\label{eq:mixed-advantage-computation}
    A_k(s,a) = \mathcal{Q}_k(s,a) - \mathcal{Q}_k(s, \pi_{\text{new}}(s))
\end{equation}
This formulation leverages the specific value estimation expertise of each critic for their respective objective.

Finally, we scalarise these vector-valued advantages into a single training signal.
We compute the \textit{mixed advantage} as the dot product of the advantage vector and the target weights derived in Step 1:
\begin{equation}
\label{eq:advantage-mixing}
    A_\text{mixed}(s,a) = \mathbf{w}_{\text{target}}^\top \cdot \mathbf{A}(s,a)
\end{equation}
This scalar value $A_{mixed}$ represents the quality of the state-action $(s,a)$ specifically regarding the desired trade-off $\mathbf{w}_{\text{target}}$.

\subsection{Step 3: Mixed Advantage Weighted Regression}
\label{step:3}
To train the offspring policy $\pi_{\text{new}}$, we employ a supervised regression objective weighted by the scalarised signal derived in Equation \ref{eq:advantage-mixing}.
Our goal is to selectively clone actions that contribute positively to the specific target trade-off $\mathbf{w}_{\text{target}}$.

For a transition $(s,a)$ we compute a regression weight $\omega(s,a)$  by applying a temperature-scaled exponential to its mixed advantage:
\begin{equation}
    \omega(s,a) = \text{min}\left( \exp\left( \frac{A_\text{mixed}(s, a)}{\beta} \right), \omega_{max} \right)
\end{equation}
where $\beta > 0$ is a temperature hyperparameter and $\omega_{max}$ is a clipping threshold to ensure numerical stability.

The fresh policy $\pi_\text{new}$ is then optimised to minimise the weighted mean squared error between its predicted action and the retrieved buffer action $a$:
\begin{equation}
    \mathcal{L}_{\text{MAPEX}} = \mathbb{E}_{(s,a)} \Big[ \omega(s,a) \cdot \| \pi_{\text{new}}(s) - a \|_2^2 \Big]
\end{equation}

Once the policy has been updated for the desired epochs, it is reinserted into the population $\Pi$ and the MAPEX procedure is executed for another iteration.

\subsection{Mitigating Out-of-Distribution Error}
While MAPEX's mixed advantage values are intuitive to compute, they expose two main sources of out-of-distribution (OOD) error: 1) when a transition sampled by one objective's specialist policy is evaluated by another objective's specialist critic, and 2) when a randomly-initialised policy's action is evaluated by specialist critics.
We mitigate these issues using the following design choices.

\subsubsection{Secondary Critics}
\label{sec:secondary-critics}

When training each specialist policy $\pi_k$, we learn not only its \emph{primary} critic for objective $k$, but also a set of \emph{secondary critics} for the remaining objectives.
All critics associated with specialist $k$ are trained on the same replay buffer $\mathcal{D}_k$ generated by $\pi_k$, so that every objective can be evaluated on data that is in-distribution for the corresponding critic.

\paragraph{Notation:}
Let $m\in\{1,\dots,N\}$ index objectives and $k\in\{1,\dots,N\}$ index specialists.
We denote by $Q^{(k)}_m(s,a)$ the action-value critic that predicts the return for objective $m$ using transitions sampled from $\mathcal{D}_k$ (i.e., collected under $\pi_k$).
Under this notation, the specialist's \emph{primary critic} is $Q^{(k)}_k$, while the \emph{secondary critics} are $\{Q^{(k)}_m\}_{m\neq k}$.

\paragraph{Procedure:}
During specialist training, only the primary critic $Q^{(k)}_k$ is used to update the policy $\pi_k$.
The secondary critics $\{Q^{(k)}_m\}_{m\neq k}$ are trained in parallel on the same buffer $\mathcal{D}_k$, but they do not contribute gradients to the policy update.
After training all specialists, we collect the resulting critics into a single family$\mathcal{Q} \triangleq \{Q^{(k)}_m\}_{k=1\ldots N,\;m=1\ldots N}.$
This construction ensures that when MAPEX samples a transition $(s,a)$ that originates from a particular specialist buffer $\mathcal{D}_k$, the evaluations $\{\mathcal{Q}^{(k)}_m(s,a)\}_{m=1}^N$ are produced by critics trained on the same state--action distribution as the sampled data.

\paragraph{Practical note:}
Secondary critics may be trained alongside primary critics with no change to the policy update rule.
If single-objective specialists have already been trained, a secondary critic $\{Q^{(k)}_m\}_{m\neq k}$ can be trained offline using transitions from $\mathcal{D}_k$ and the corresponding objective rewards.
Either way, a distribution-matched critic family $\mathcal{Q}$ can be learnt for a low cost.
We include a comparison of joint-vs.\ post-hoc-trained secondary critics in Section \ref{sec:experiments}.

\subsubsection{Offspring Policy Warm-Up}
\label{sec:policy-init}

In step 2 of the MAPEX procedure (Equation \ref{eq:mixed-advantage-computation}), computing the mixed advantage requires querying specialist critics with actions proposed by $\pi_{\text{new}}(s)$.
If $\pi_{\text{new}}$ is initialised arbitrarily, these actions can be far from the support of the hybrid buffer and cause OOD error.
To reduce this effect, we warm up $\pi_{\text{new}}$ by regressing it to the mean of its parents in the action space.

Let $\{\pi_{p_1},\dots,\pi_{p_N}\}$ denote the parent policies.
The mean parent action at state $s$ is
$\bar a(s) \;\triangleq\; \frac{1}{N}\sum_{j=1}^{N}\,\pi_{p_j}(s)$,
which we use to perform a brief behavioural regression step that minimises
\begin{equation}
L_{\text{init}}(\theta) \;\triangleq\;
\mathbb{E}_{s\sim \mathcal{D}_{\text{hybrid}}}
\left[\left\|\pi_{\text{new}}(s)-\bar a(s)\right\|_2^2\right],
\label{eq:policy-init-loss}
\end{equation}
We run this regression for a small number of gradient steps prior to computing the advantage vector in Equation \ref{eq:mixed-advantage-computation}.
This warm up keeps $\pi_{new}$'s predicted actions close to those produced by the parents on states drawn from $\mathcal{D}_{\text{hybrid}}$, more closely matching each critic's training exposure, and improving the reliability of subsequent critic-based updates.

\section{Experiments}
\label{sec:experiments}
We evaluate MAPEX on three criteria: \textbf{sample efficiency} of Pareto extraction; \textbf{flexibility} regarding choice of specialist training algorithm and nature of secondary critic training (joint vs.\ post-hoc); and \textbf{general competitiveness} against baselines when training from scratch. 

We assume access to secondary critics trained alongside primary critics during specialist training, but also test a \textit{post hoc} variant (MAPEX-PostHoc) where critics are trained offline on static buffers. Unless stated otherwise, specialists are trained using Proximal Distilled Evolutionary RL (PDERL) \cite{pderl}.

\subsection{Baselines and Domains}
We compare MAPEX against two established MORL methods:
\begin{itemize}
    \item \textbf{MOPDERL} \cite{mopderl}: An evolutionary actor-critic method that first trains on each individual objective using PDERL, and later crosses over solutions across objectives using a multi-objective distilled crossover. It serves as a baseline for both `from scratch' training and pure Pareto extraction (via its distillation phase).
    \item \textbf{MORL/D} \cite{morld}: A decomposition-based approach that uses Soft Actor--Critic \cite{soft-actor-critic} on each scalar objective. It uses buffer data sharing and adapts the scalar weights using Pareto Simulated Annealing \cite{pareto-simualated-annealing}.
\end{itemize}
We use the \texttt{morl-baselines} \cite{mo-gymnasium} implementation of MORL/D and the authors' original implementation of MOPDERL \footnote{Modified for v5 MO-Gymnasium MuJoCo domains. Code: https://github.com/raghavthakar/MOPDERL-MO-Gymnasium} with default hyperparameters. Experiments are conducted on five bi-objective continuous control MuJoCo environments from \texttt{MO-Gymnasium} \cite{mo-gymnasium}, with episodes capped at 750 frames.
The nature of objectives in each environment are specified in Appendix \ref{sec:env-details} table \ref{tab:envs}.

\subsection{Experimental Setup}
\textbf{Sample efficiency:} We compare the extraction phase of MAPEX against the distillation phase of MOPDERL starting from identical specialist subpopulations trained with PDERL.
\textbf{Flexibility:} We compare standard MAPEX against MAPEX-PostHoc and MAPEX-TD3 (specialists trained via TD3) to assess robustness to starting policies and nature of critic training.
\textbf{Competitiveness:} We compare the final Pareto fronts of the full MAPEX pipeline against MOPDERL and MORL/D given a fixed sample budget.
During specialist training of all MAPEX variants, we use a replay buffer of size 1M.
In each test we perform five seeded runs with each method.
Exact algorithm-specific and experiment parameters are listed in Appendix \ref{sec:hyperparams}.

\section{Results}
\label{sec:results}

\subsection{Sample Efficiency of Extraction}
\begin{figure*}[t!]
    \centering
    \includegraphics[width=1\linewidth,trim={0 0.2cm 0 0},clip]{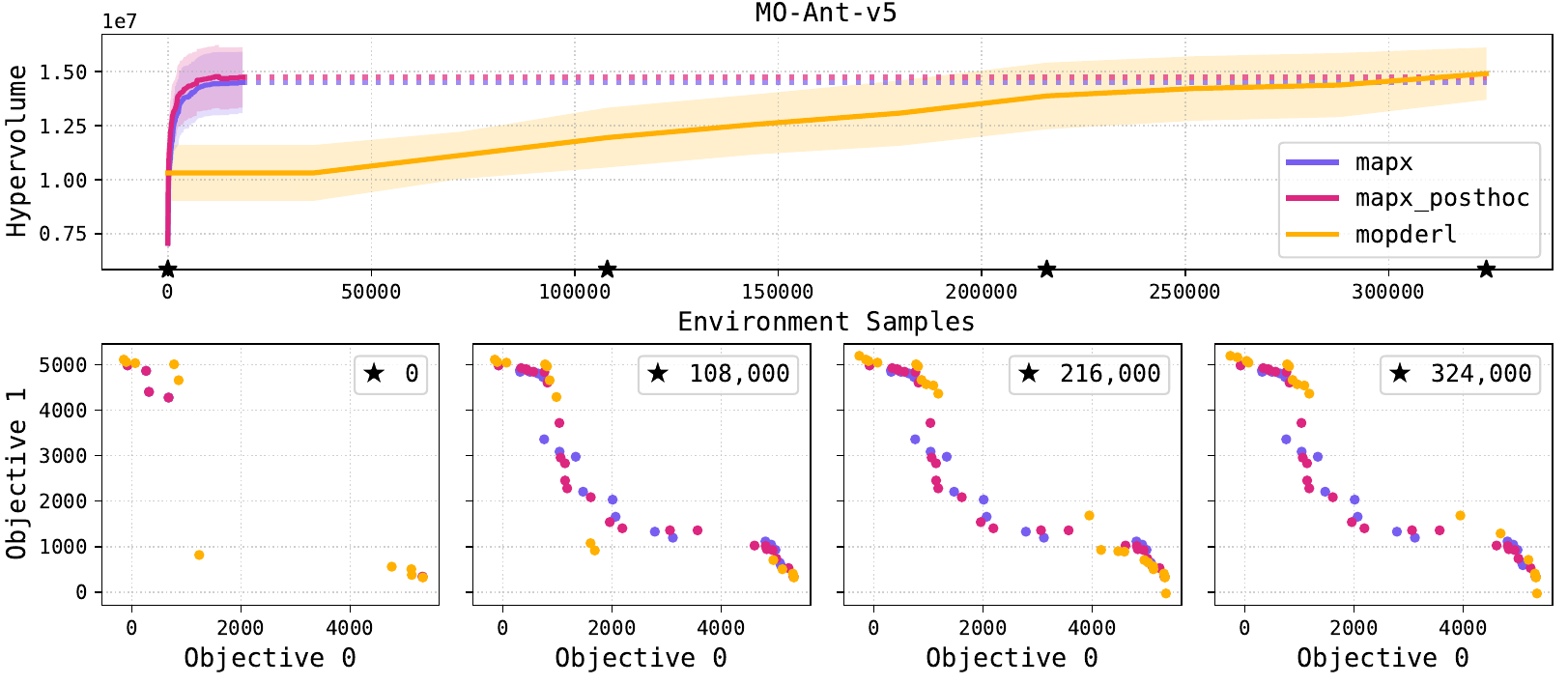}
    \caption{\textbf{Sample efficiency comparison on MO-Ant-v5.} (Top) Mean hypervolume $\pm$ SEM vs. cumulative environment samples. MAPEX and MAPEX-PostHoc achieve high hypervolume almost instantaneously, while MOPDERL requires significantly more interaction. (Bottom) Evolution of the Pareto front approximation. MAPEX/MAPEX-PostHoc fill the front immediately, whereas MOPDERL gradually expands coverage over 300,000+ environment samples.}
    \label{fig:ant-composite}
\end{figure*}

MAPEX achieves a massive reduction in sample cost compared to baselines. As shown in Figure \ref{fig:ant-composite} (MO-Ant-v5), MAPEX and MAPEX-PostHoc extract a high-performing front almost immediately, while MOPDERL requires an additional $300,000$ environment interactions to attain the same performance.

\begin{figure}[h!]
    \centering
    \includegraphics[width=1\linewidth]{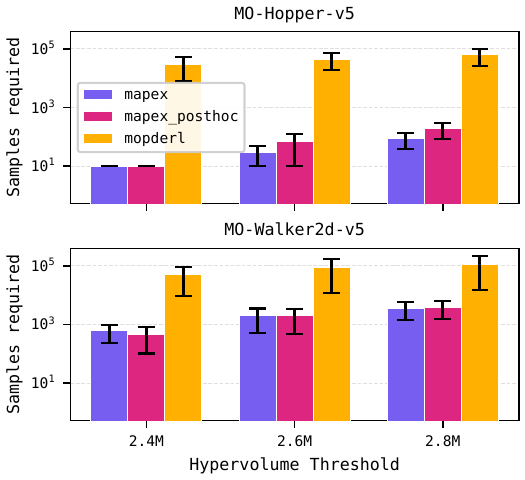}
    \caption{\textbf{Samples required to attain target hypervolume thresholds.} Comparison on MO-Hopper-v5 (top) and MO-Walker2d-v5 (bottom). Note the logarithmic scale on the y-axis. MAPEX and MAPEX-PostHoc require up to three orders of magnitude fewer samples ($10^2$ vs $10^5$) than MOPDERL to reach identical performance levels.}
    \label{fig:hopper-walker-bar}
\end{figure}

Figure \ref{fig:hopper-walker-bar} confirms this trend across MO-Hopper-v5 and MO-Walker2d-v5. Particularly in MO-Hopper-v5, MAPEX requires $100$ samples to reach hypervolume thresholds that MOPDERL requires $\approx10^5$ samples to attain---a reduction of three orders of magnitude. This efficiency is driven by MAPEX exploiting the latent representations of multi-objective behaviour in policies and replay buffers.
It empirically validates our intuition that following expert behaviour to varying degrees on each objective yields trade-offs in the objective space. It also validates our main hypothesis that these behaviours can be learnt by regressing onto target actions, weighed by their \textit{mixed advantage} value.

\begin{figure}[h!]
    \centering
    \includegraphics[width=1\linewidth]{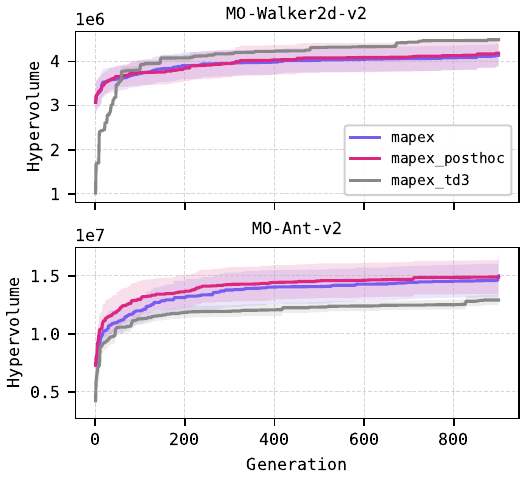}
    \caption{\textbf{Robustness of MAPEX to specialist type and critic training.} Mean hypervolume (± SEM) over generations on MO-Walker2d-v5 and MO-Ant-v5. The similar performance of standard MAPEX, MAPEX-PostHoc (offline critics), and MAPEX-TD3 (off-policy specialists) demonstrates the method's flexibility in effectively extracting fronts from  decoupled pre-trained sources.}
    \label{fig:walker-ant-flexibility}
\end{figure}

\subsection{Flexibility and Robustness}
\begin{table*}[h!]
\centering
\caption{Performance metrics (Hypervolume and Sparsity) across v5 \texttt{MO-Gymnasium} MuJoCo environments. Results are Mean $\pm$ SEM. Hypervolume is the space between the Pareto front and a dominated reference point and sparsity is the average euclidean distance between neighbouring points.}
\label{tab:metrics_compact}
\small 
\setlength{\tabcolsep}{0pt} 
\begin{tabular*}{\textwidth}{@{\extracolsep{\fill}}l ccc ccc @{}}
\toprule
 & \multicolumn{3}{c}{\textbf{Hypervolume} ($\uparrow$)} & \multicolumn{3}{c}{\textbf{Sparsity} ($\downarrow$)} \\
\cmidrule(lr){2-4} \cmidrule(l){5-7}
\textbf{Environment} & \textbf{MAPX} & \textbf{MOPDERL} & \textbf{MORL/D} & \textbf{MAPEX} & \textbf{MOPDERL} & \textbf{MORL/D} \\
\midrule
Ant-2obj 
 & $1.19\text{e}7 \pm 1.2\text{e}6$ & $1.46\text{e}7 \pm 8.2\text{e}5$ & $1.41\text{e}7 \pm 1.9\text{e}4$ 
 & $315.5 \pm 36.1$ & $\mathbf{201.4 \pm 13.8}$ & $289.8 \pm 8.3$ \\
Hopper-2obj 
 & $3.34\text{e}6 \pm 2.9\text{e}5$ & $3.17\text{e}6 \pm 2.8\text{e}5$ & $\mathbf{4.31\text{e}6 \pm 1.7\text{e}5}$ 
 & $209.6 \pm 39.8$ & $\mathbf{64.2 \pm 10.3}$  & $115.5 \pm 17.7$ \\
Swimmer 
 & $1.78\text{e}5 \pm 1.6\text{e}4$ & $\mathbf{2.41\text{e}5 \pm 2.9\text{e}4}$ & $9.29\text{e}4 \pm 5.9\text{e}2$ 
 & $49.5 \pm 9.2$   & $30.0 \pm 3.2$   & $\mathbf{2.9 \pm 0.3}$ \\
Walker2d 
 & $3.09\text{e}6 \pm 2.2\text{e}5$ & $3.47\text{e}6 \pm 3.5\text{e}5$ & $\mathbf{6.52\text{e}6 \pm 4.2\text{e}5}$ 
 & $157.5 \pm 7.9$  & $103.2 \pm 16.0$ & $96.9 \pm 11.2$ \\
HalfCheetah 
 & $1.10\text{e}7 \pm 5.3\text{e}5$ & $1.55\text{e}7 \pm 6.8\text{e}5$ & $\mathbf{1.88\text{e}7 \pm 5.4\text{e}4}$ 
 & $337.5 \pm 26.7$ & $\mathbf{101.7 \pm 5.9}$  & $136.4 \pm 19.6$ \\
\bottomrule
\end{tabular*}
\end{table*}

\begin{figure*}
    \includegraphics[width=1\linewidth,trim={0 0.9cm 0 0},clip]{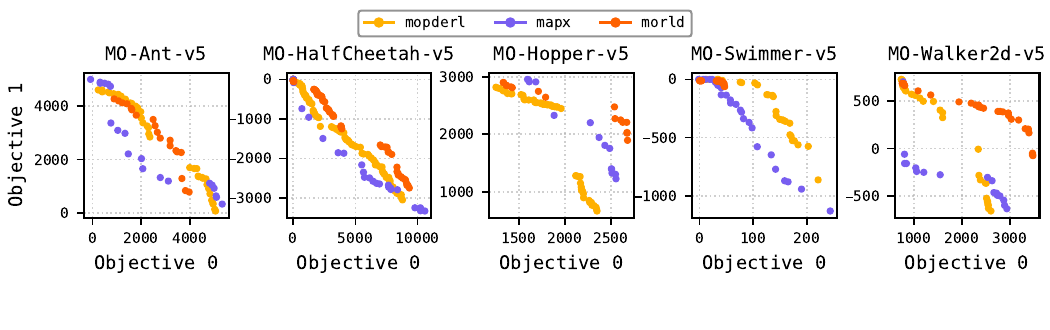}
    \caption{\textbf{Final Pareto fronts across five MO-MuJoCo benchmarks.} Comparison of fronts extracted by MAPEX against fully trained MOPDERL and MORL/D baselines. Despite relying purely on single-objective training data, MAPEX recovers fronts that are dense and competitive with baselines trained from scratch.}
    \label{fig:all-fronts}
\end{figure*}
MAPEX is robust to the source of specialist policies. Figure \ref{fig:walker-ant-flexibility} shows that Pareto extraction for standard MAPEX, MAPEX-PostHoc, and MAPEX-TD3 are largely indistinguishable on MO-Walker2d-v5 and MO-Ant-v5. Crucially, the success of MAPEX-PostHoc confirms that secondary critics can be effectively trained retroactively on static buffers, enabling Pareto extraction from fully decoupled single-objective training. If simplicity is key, then the performance of MAPEX-TD3 shows the potential of elegantly adapting off-policy RL to learn multi-objective behaviours via MAPEX.

While a population of policies (like that produced by PDERL) provides a rich starting point for Pareto extraction, MAPEX-TD3 remains competitive with only one specialist policy per objective (as produced by simple TD3).
This is because MAPEX does not simply distil or interpolate between parent policies.
Instead, MAPEX draws expertise from the pre-trained replay buffers, which contain a rich and varied set of state-action examples.
These experiences span regions of the objective space that an individual policy may only cover sparsely, allowing MAPEX to produce distinct policies for distinct trade-offs.
This is also why MAPEX is robust to the exact algorithm used to train specialist policies.

\subsection{General Competitiveness}
Despite being an offline extraction method, MAPEX produces fronts competitive with MOPDERL and MORL/D, which require full multi-objective consideration from the beginning. Table \ref{tab:metrics_compact} shows that MAPEX achieves comparable hypervolumes (e.g., $3.34\times10^6$ vs.\ MOPDERL's $3.17\times10^6$ in MO-Hopper-v5, and $1.78\times10^5$ vs.\ MORL/D's $9.29\times10^4$ in MO-Swimmer-v5) across five multi-objective MuJoCo environments.
While MAPEX fronts are numerically sparser, by inspecting the fronts in Figure \ref{fig:all-fronts} visually, it is clear that MAPEX produces even and well-spread Pareto fronts.

\section{Conclusion}
We presented MAPEX, a novel approach to extracting Pareto fronts of policies from prior single-objective training for continuous control.
We provided a detailed view of MAPEX's Pareto extraction procedure, and mentioned some practial tips.
We tested MAPEX with well-established, dedicated MORL baselines like MOPDERL and MORL/D to empirically validate its mixed advantage approach.
Pareto fronts are learnt cheaply with MAPEX when specialists are already trained.
If training from scratch, MAPEX integrates easily with off-policy RL methods and still produces fronts that compare well to baselines. Finally, we discuss some limitations and future work.

While MAPEX is highly sample efficient, it makes assumptions inherent to our offline extraction setting.
First, MAPEX is strictly bounded by the support of the specialist buffers; it cannot discover novel behaviours or skills that are absent from the specialists' training history.
Second, MAPEX relies on the assumption that valid trade-off policies lie on a continuous manifold between specialists.
In scenarios where specialists exhibit markedly disjoint behaviours (e.g., a humanoid walking on two legs vs. crawling), interpolation may yield low-performance policies.
Our empirical evaluation focused on bi-objective domains; scaling MAPEX's gap-identification heuristic to more objectives ($N \ge 3$) remains a subject for future investigation.
Finally, we would like to leverage MAPEX with a multiagent reinforcement learning algorithm to extend multi-objective decision-making to the multiagent setting.

\bibliographystyle{named}
\bibliography{ijcai26}

\appendix
\section{Environment Details}
\label{sec:env-details}
Table \ref{tab:envs} specifies each objective across five multi-objective MuJoCo environments.
Episodes in each environment are capped at 750 environment steps.

\begin{table}
\centering
\caption{The v5 \texttt{MO-Gymnasium} MuJoCo environments. Objectives include velocity, energy efficiency, jump height, and stability and must all be \emph{maximised}. The action dimensions are $d_a$, and $d_o$ the observation (state) dimensions.}
\label{tab:envs}
\small
\setlength{\tabcolsep}{3pt} 
\begin{tabular}{@{}lccp{2.3cm}p{2.3cm}@{}}
\toprule
Env. & $d_a$ & $d_o$ & Objective 1 & Objective 2 \\
\midrule
Ant-2obj & 8 & 105 & x-vel & y-vel \\
Hopper-2obj & 3 & 11 & x-vel \& survival & Jump (w/ cost) \\
Swimmer & 2 & 8 & x-vel & Energy eff. \\
Walker2d & 6 & 17 & x-vel & Energy eff. \\
HalfCheetah & 6 & 17 & x-vel & Energy eff. \\
\bottomrule
\end{tabular}
\end{table}
\section{Experimental Hyperparameters}
\label{sec:hyperparams}

We report the hyperparameters used in our experiments. Table~\ref{tab:pderl_td3_params} lists the settings for the base TD3 algorithm and the PDERL algorithm which uses TD3, which are shared across all environments. Table~\ref{tab:mapex-global} details the general parameters for MAPEX, and Table~\ref{tab:mapex-env} lists the environment-specific values for policy warm up and Pareto extraction.
Finally Table \ref{tab:frame_budgets} lists the number of frames each algorithm was run for, and the division of frames for MOPDERL, which contains distinct warm up and Pareto distillation phases.

\begin{table}[h]
\centering
\caption{PDERL and TD3 hyperparameters (Shared across environments).}
\label{tab:pderl_td3_params}
\small
\begin{tabular}{@{}ll@{}}
\toprule
\textbf{Parameter} & \textbf{Value} \\
\midrule
\multicolumn{2}{l}{\textit{PDERL}} \\
Population Size & 10 \\
Mini Buffer Size & 50,000 \\
\midrule
\multicolumn{2}{l}{\textit{TD3 / General}} \\
Start Timesteps & $25,000$ \\
Discount Factor ($\gamma$) & 0.99 \\
Target Smoothing ($\tau$) & 0.005 \\
Hidden Dimension & 256 \\
Actor Learning Rate & $3 \times 10^{-4}$ \\
Critic Learning Rate & $3 \times 10^{-4}$ \\
Batch Size & 256 \\
Buffer Size & $1 \times 10^6$ \\
Exploration Noise & 0.1 \\
Policy Noise & 0.2 \\
Noise Clip & 0.5 \\
Policy Frequency & 2 \\
\bottomrule
\end{tabular}
\end{table}

\begin{table}[h]
\centering
\caption{MAPEX hyperparameters shared across environments.}
\label{tab:mapex-global}
\begin{tabular}{@{}ll@{}}
\toprule
\textbf{Hyperparameter} & \textbf{Value} \\
\midrule
Total iterations & 1,200 \\
Child buffer size & 200{,}000 \\
Warm-up steps / epoch & 1{,}000 \\
Warm-up learning rate & $3\times 10^{-4}$ \\
Warm-up batch size & 256 \\
Evaluation episodes / actor & 5 \\
\bottomrule
\end{tabular}
\end{table}

\begin{table*}
\centering
\caption{Environment-specific MAPEX hyperparameters across five v5 \texttt{MO-Gymnasium} MuJoCo environments.}
\label{tab:mapex-env}
\begin{tabular}{lccccc}
\toprule
\textbf{Parameter} & \multicolumn{5}{c}{\textbf{Environment}} \\
\cmidrule(lr){2-6}
 & \textbf{Ant-2obj} & \textbf{Hopper-2obj} & \textbf{Swimmer} & \textbf{Walker2d} & \textbf{HalfCheetah} \\
\midrule
MAPEX epochs & 20 & 10 & 10 & 20 & 20 \\
AWR $\beta$ & 0.5 & 0.1 & 1.0 & 0.5 & 1.0 \\
AWR clip ($\omega_{\max}$) & 1.0 & 20.0 & 20.0 & 20.0 & 1.0 \\
\bottomrule
\end{tabular}
\end{table*}

\begin{table*}
\centering
\caption{Environment interaction budgets (frames) per method and environment. For bi-objective tasks ($N{=}2$), the “/ obj.” column reports frames per objective-specific training process (PDERL for MOPDERL; PDERL / TD3 for MAPEX). MAPEX extraction uses 0 additional environment interaction.}
\label{tab:frame_budgets}
\small
\setlength{\tabcolsep}{4pt}
\begin{tabular}{lccccccc}
\toprule
\multirow{2}{*}{\textbf{Environment}} &
\multicolumn{4}{c}{\textbf{MOPDERL}} &
\multicolumn{2}{c}{\textbf{MAPEX (specialists)}} &
\multirow{2}{*}{\textbf{MORL/D} \textbf{Total}} \\
\cmidrule(lr){2-5}\cmidrule(lr){6-7}
& \textbf{Warm up} & \textbf{Warm up / obj.} & \textbf{Stage 2} & \textbf{Total} &
\textbf{Specialist / obj.} & \textbf{Total} & \\
\midrule
MO-Ant-2obj-v5         & 2.0M & 1.0M   & 2.0M  & 4.0M  & 2.0M & 4.0M & 4.0M \\
MO-Hopper-2obj-v5      &2.25M & 1.125M & 1.75M & 4.0M  & 2.0M & 4.0M & 4.0M \\
MO-Swimmer-v5     & 1.0M & 0.5M   & 1.0M  & 2.0M  & 1.0M & 2.0M & 2.0M \\
MO-Walker2d-v5    & 2.0M & 1.0M   & 2.0M  & 4.0M & 2.0M & 4.0M & 4.0M \\
MO-HalfCheetah-v5 & 2.0M & 1.0M   & 2.0M  & 4.0M  & 2.0M & 4.0M & 4.0M \\
\bottomrule
\end{tabular}
\end{table*}

\end{document}